\documentclass{edm_article}
\usepackage{comment}

\usepackage{graphicx}

\usepackage{wrapfig}
\usepackage{makecell}
\usepackage{multirow}
\usepackage{subfigure}

\usepackage{tikz}
\usepackage{amsmath,amssymb} 
\DeclareMathOperator*{\argmax}{argmax}

\usepackage{url}

\begin{document}


\title{Towards Explainable Student Group Collaboration Assessment Models Using Temporal Representations of \\Individual Student Roles}


%
\numberofauthors{6}
\author{
\alignauthor
 Anirudh Som\\
        \affaddr{Center for Vision Technologies}\\
        \affaddr{SRI International}\\
        \email{anirudh.som@sri.com}
 \alignauthor
 Sujeong Kim\\
        \affaddr{Center for Vision Technologies}\\
        \affaddr{SRI International}\\
        \email{sujeong.kim@sri.com}
 \alignauthor 
 Bladimir Lopez-Prado\\
        \affaddr{Center for Education Research and Innovation}\\
        \affaddr{SRI International}\\
        \email{bladimir.lopez-prado@sri.com}
 \and  
 \alignauthor Svati Dhamija\\
        \affaddr{Center for Vision Technologies}\\
        \affaddr{SRI International}\\
        \email{svati.dhamija@sri.com}
 \alignauthor Nonye Alozie\\
        \affaddr{Center for Education Research and Innovation}\\
        \affaddr{SRI International}\\
        \email{maggie.alozie@sri.com}
 \alignauthor Amir Tamrakar\\
        \affaddr{Center for Vision Technologies}\\
        \affaddr{SRI International}\\
        \email{amir.tamrakar@sri.com}
 }

\maketitle


\begin{abstract}
Collaboration is identified as a required and necessary skill for students to be successful in the fields of Science, Technology, Engineering and Mathematics (STEM). However, due to growing student population and limited teaching staff it is difficult for teachers to provide constructive feedback and instill collaborative skills using instructional methods. Development of simple and easily explainable machine-learning-based automated systems can help address this problem. Improving upon our previous work, in this paper we propose using simple temporal-CNN deep-learning models to assess student group collaboration that take in temporal representations of individual student roles as input. We check the applicability of dynamically changing feature representations for student group collaboration assessment and how they impact the overall performance. We also use Grad-CAM visualizations to better understand and interpret the important temporal indices that led to the deep-learning model's decision. 
\end{abstract}

%


\section{Introduction}

Collaboration is considered a crucial skill, that needs to be inculcated in students early on for them to excel in STEM fields \cite{ngss2013next,daggett2010common}. Traditional instruction-based methods \cite{krajcik2006project,davidson2014boundary} can often make it difficult for teachers to observe several student groups and identify specific behavioral cues that contribute or detract from the collaboration effort \cite{smith2008guided,loughry2007development,taggar2001problem}. This has resulted in a surge in interest to develop machine-learning-based automated systems to assess student group collaboration \cite{reilly2019predicting, huang2019identifying, kang2019collaborative, genolini2011kml, alexandron2020towards, guo2019collaboration, talavera2004mining, spikol2017using,soller2002machine,anaya2011application,vrzakova2020focused,som2020machine}.

In our earlier work we developed a multi-level, multi-modal conceptual model that serves as an assessment tool for individual student behavior and group-level collaboration quality \cite{alozie2020automated,alozie2020collaboration}. Using the conceptual model as a reference, in a different paper we developed simple MLP deep-learning models that predict student group collaboration quality from histogram representations of individual student roles \cite{som2020machine}. Please refer to the following papers for more information and for the illustration of the conceptual model \cite{alozie2020automated,alozie2020collaboration,som2020machine}. Despite their simplicity and effectiveness, the MLP models and histogram representations lack explainability and insight into the important student dynamics. To address this, in this paper we focus on using simple temporal-CNN deep learning models to check the scope of dynamically changing temporal representations for student group collaboration assessment. We also use Grad-CAM visualizations to help identify important temporal instances of the task performed and how they contribute towards the model's decision.

\textbf{Paper Outline:} Section \ref{section_background} provides necessary background on the different loss functions used, dataset description and the temporal features extracted. Section \ref{section_experiments} describes the experiments and results. Section \ref{section_conclusion} concludes the paper.

\section{Background}\label{section_background}


\subsection{Cross-Entropy Loss Functions}

The categorical-cross-entropy loss is the most commonly used loss function to train deep-learning models. For a classification problem with $C$ classes, let us denote the input variables as $\mathbf{x}$, ground-truth label vector as $\mathbf{y}$ and the predicted probability distribution as $\mathbf{p}$. Given a training sample $(\mathbf{x},\mathbf{y})$, the categorical-cross-entropy (CE) loss is defined as 

\begin{equation}\label{equation_cross-entropy}
    \text{CE}_{\mathbf{x}}(\mathbf{p},\mathbf{y}) = - \sum_{i=1}^{C} \mathbf{y}_i\log(\mathbf{p}_i)
\end{equation}

Here, $\mathbf{p}_i$ denotes the predicted probability of the $i$-th class. Note, both $\mathbf{y}$ and $\mathbf{p}$ are of length $C$, with $\sum_i \mathbf{y}_i = \sum_i \mathbf{p}_i = 1$. From Equation \ref{equation_cross-entropy}, it's clear that for imbalanced datasets the learnt weights of the model will be biased towards classes with the most number of samples in the training set. Additionally, if the label space exhibits an ordered structure, the categorical-cross-entropy loss will only focus on the predicted probability of the ground-truth class while ignoring how far off the incorrectly predicted sample actually is. These limitations can be addressed to some extent by using the ordinal-cross-entropy (OCE) loss function \cite{som2020machine}, defined in Equation \ref{equation_ordinal-cross-entropy}.

\begin{equation}\label{equation_ordinal-cross-entropy}
\begin{split}
    \text{OCE}_{\mathbf{x}}(\mathbf{p},\mathbf{y}) &= - \left(1+w\right)\sum_{i=1}^{C} \mathbf{y}_i\log(\mathbf{p}_i) \\ \text{ } w &= \left|\argmax(\mathbf{y}) - \argmax(\mathbf{p}) \right|
\end{split}
\end{equation}

\begin{table}[t!]
	\centering
	\caption{Coding rubric for Level A and Level B2.}\label{table_level-a-b2-codes}
	\scalebox{0.82}{
	\begin{tabular}{ |c|c| } 
			\hline
			\textbf{Level A} & \textbf{Level B2}\\ 
			\hline
			 & Group guide/Coordinator [GG] \\
			Effective [E] &	Contributor (Active) [C] \\
			Satisfactory [S]&	Follower [F]\\
			Progressing [P] &	Conflict Resolver [CR] \\
		    Needs Improvement [NI] &	Conflict Instigator/Disagreeable [CI] \\
			Working Independently [WI] 	&	Off-task/Disinterested [OT] \\
			&	Lone Solver [LS] \\
			\hline
	\end{tabular}}
\end{table}

Here, $(1+w)$ represents the weighting variable, argmax returns the index of the maximum valued element and $|.|$ returns the absolute value. When training the model, $w = 0$ for correctly classified training samples, with the ordinal-cross-entropy loss behaving exactly like the categorical-cross-entropy loss. However, for misclassified samples the ordinal-cross-entropy loss will return a higher loss value. The increase in loss is proportional to how far away a sample is misclassified from its ground-truth class label. 

\begin{table}[tb!]
	\centering
	\caption{Inter-rater reliability (IRR) measurements.}\label{table_IRR}
	\scalebox{0.99}{
		\begin{tabular}{ |c|c|c| } 
			\hline
			\textbf{Level} & \textbf{Average Agreement} & \textbf{Cohen's Kappa} \\ 
			\hline
		    A & 0.7046 & 0.4908 \\
            \hline
            B2 & 0.6741 & 0.5459 \\
            \hline
	\end{tabular}}
\end{table}

\subsection{Dataset Description}\label{subsection_dataset_description}

We collected audio and video recordings from 15 student groups, across five middle schools. Out of the 15 groups, 13 groups had 4 students, 1 group had 3 students, and 1 group had 5 students. The student volunteers completed a brief survey that collected their demographic information and other details, e.g., languages spoken, ethnicity and comfort levels with science concepts. Each group was tasked with completing 12 open-ended life science and physical science tasks, which required them to construct models of different science phenomena as a team. They were given one hour to complete as many tasks possible, which resulted in 15 hours of audio and video recordings. They were provided logistic and organization instructions but received no help in group dynamics, group organization, or task completion. 

Next, the data recordings were manually annotated by education researchers at SRI International. For the rest of the paper we will refer to them as coders/annotators. In our hierarchical conceptual model \cite{alozie2020automated,alozie2020collaboration}, we refer to the collaboration quality annotations as Level A and individual student role annotations as Level B2. The coding rubric for these two levels is described in Table \ref{table_level-a-b2-codes}. Both levels were coded by three annotators. They had access to both audio and video recordings and used ELAN (an open-source annotation software) to annotate. A total of 117 tasks were coded by each annotator, with the duration of each task ranging from 5 to 24 minutes. Moderate-agreement was observed across the coders as seen from the inter-rater reliability measurements in Table \ref{table_IRR}.


Level A codes represent the target label categories for our classification problem. To determine the ground-truth Level A code, the majority vote (code) across the three annotators was used as the ground-truth. For cases where a majority was not possible, we used the Level A code ordering depicted in Table \ref{table_level-a-b2-codes} to determine the median as ground-truth of the three codes. For example, if the three coders assigned \textit{Satisfactory, Progressing, Needs Improvement} for the same task then \textit{Progressing} would be used as the ground-truth label. Note, we did not observe a majority Level A code for only 2 tasks. To train the machine learning models  we only had 351 data samples (117 tasks $\times$ 3 coders).

\begin{figure}[tb!]
\centering
\includegraphics[width=0.99\linewidth]{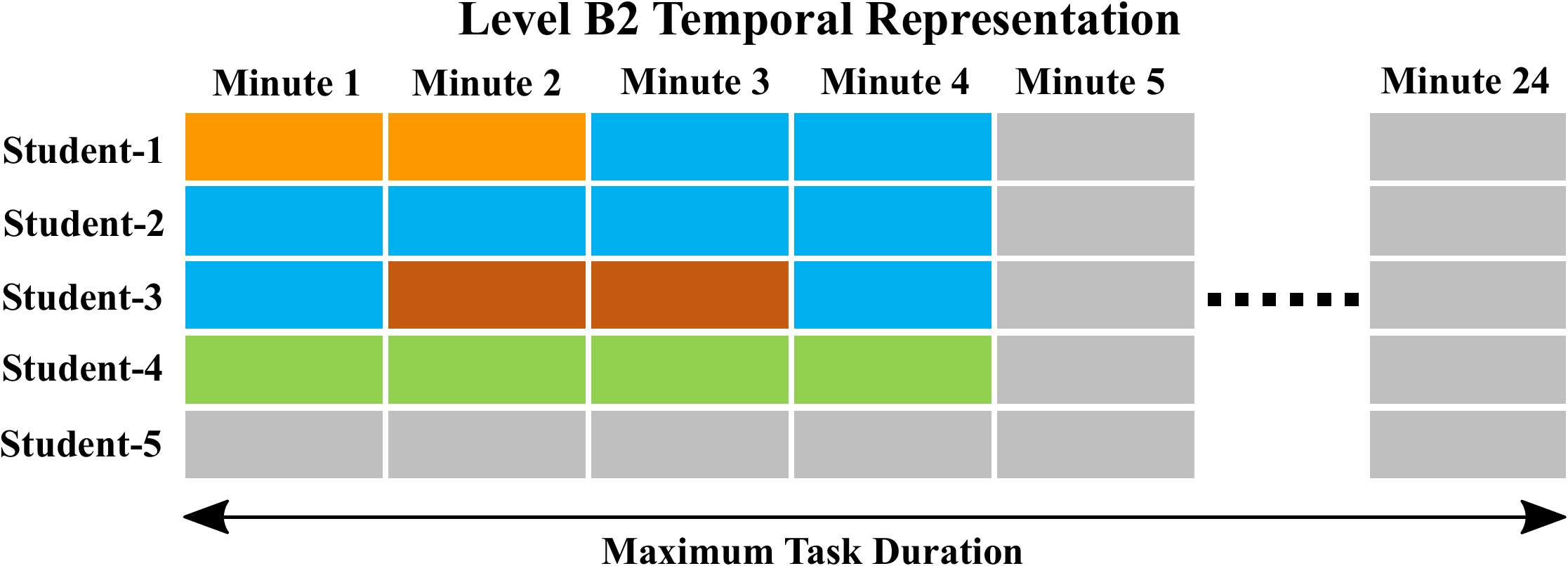}
	\caption{Level B2 temporal representation for a group having only 4 students and finishing the assigned task in 4 minutes. Colored cells illustrate the different Level B2 codes as described in Table \ref{table_level-a-b2-codes}, and the gray cells represent empty or unassigned codes.}\label{figure_temp_b2}
\end{figure}

\subsection{Temporal Representation}\label{subsection_temporal-representations}

In our dataset, the longest task was little less than 24 minutes, due to which the length for all tasks was also set to 24 minutes. Level B2 was coded using fixed-length 1 minute segments, as illustrated in Figure \ref{figure_temp_b2}. Due to its fixed-length nature, we assigned an integer value to each B2 code, i.e., the seven B2 codes were assigned values from 1 to 7. The value 0 was used to represent segments that were not assigned a code. For example, in Figure \ref{figure_temp_b2} we see a group of 4 students completing a task in just 4 minutes, represented by the colored cells. The remaining 20 minutes and the 5$^\text{th}$ student is assigned a value zero, represented by the gray cells. Thus for each task, Level B2 temporal features will have a shape $24\times 5$, with 24 representing number of minutes and 5 representing number of students in the group.

\textbf{Baseline Histogram Representation:} We compare the performance of the temporal representations against simple histogram representations \cite{som2020machine}. The histogram representations were created by pooling over all the codes observed over the duration of the task and across all the students. Note, only one histogram was generated per task, per group. Once the histogram is generated we normalize it by dividing by the total number of codes in the histogram. Normalizing the histogram removes the temporal aspect of the task. For example, if group-1 took 10 minutes to solve a task and group-2 took 30 minutes to solve the same task, but both groups were assigned the same Level A code despite group-1 finishing the task sooner. The raw histogram representations of both these groups would look different due to the difference in number of segments coded. However, normalized histograms would make them more comparable. Despite the normalized histogram representation being simple and effective, it fails to offer any insight or explainability. 

\section{Experiments}\label{section_experiments}

\textbf{Network Architecture:} For the temporal-CNN deep learning model we used the temporal ResNet architecture described in \cite{wang2017time}. The ResNet architecture uses skip connections between each residual block to help avoid the vanishing gradient problem. It has shown state-of-the-art performance in several computer vision applications \cite{he2016deep}. Following \cite{wang2017time}, our ResNet model consists of three residual blocks stacked over one another, followed by a global-average-pooling layer and a softmax layer. The number of filters for each residual block was set to 64, 128, 128 respectively. The number of learnable parameters for the B2 temporal representations is 506949. We compare the performance of the ResNet model to the MLP models described in our previous work. Interested readers should refer to \cite{som2020machine} for more information about the baseline MLP model that was used with the histogram representation. 

\textbf{Training and Evaluation Protocol:} All models were developed using Keras with TensorFlow backend \cite{chollet2015keras}. We used the Adam optimizer \cite{kingma2014adam} and trained all models for 500 epochs. The batch-size was set to one-tenth of the number of training samples during any given training-test split. We optimized over the Patience and Minimum-Learning-Rate hyperparameters, that were set during the training process. We focused on these as they significantly influenced the model's classification performance. The learning-rate was reduced by a factor of 0.5 if the loss did not change after a certain number of epochs, indicated by the Patience hyperparameter. We saved the best model that gave us the lowest test-loss for each training-test split. We used a round-robin leave-one-group-out cross validation protocol. This means that for our dataset consisting of $g$ student groups, for each training-test split we used data from $g-1$ groups for training and the left-out group was used as the test set. This was repeated for all $g$ groups and the average result was reported. For our experiments $g=14$ though we have temporal representations from 15 student groups. This is because all samples corresponding to the \emph{Effective} class were found only in one group. Due to this reason and because of our cross-validation protocol we do not see any test samples for the \emph{Effective} class.

\subsection{Temporal vs Histogram Representations}

Here, we compare the performance of the ResNet and MLP models. Using the weighted F1-score performance, Table \ref{table_best-results} summarizes the best performing ResNet and MLP models for the different feature-classifier variations. The table also provides the weighted precision and recall metrics. Bold values in the table represent the best classifier across the different feature-classifier settings. The ordinal-cross-entropy loss with or without class-balancing shows the highest weighted F1-score performance for both feature types. Here, class-balancing refers to weighting each data sample by a weight that is inversely proportional to the number of data samples corresponding to that sample's ground-truth label.

\begin{table}[tb!]

	\centering
	\caption{Weighted precision, weighted recall and weighted F1-score Mean$\pm$Std for the best MLP and ResNet models under different settings.}\label{table_best-results}
	\scalebox{0.6}{
		\begin{tabular}{ |c|c|c|c|c| } 
			\hline
			\textbf{Feature} & \textbf{Classifier} & \textbf{\makecell{Weighted\\Precision}} & \textbf{\makecell{Weighted\\Recall}} & \textbf{\makecell{Weighted\\F1-Score}}\\
			\hline
			\hline
			\multirow{7}{*}{\makecell{B2\\ Histogram}} & SVM & 84.45$\pm$13.43 & 73.19$\pm$16.65 & 76.92$\pm$15.39 \\ 
			& MLP - Cross-Entropy Loss & 83.72$\pm$16.50 & 86.42$\pm$10.44 & 84.40$\pm$13.85 \\
			& \makecell{MLP - Cross-Entropy Loss \\ + Class-Balancing} & 83.93$\pm$17.89 & 85.29$\pm$14.37 & 84.16$\pm$16.23 \\
			& MLP - Ordinal-Cross-Entropy Loss & 86.96$\pm$14.56 & 88.78$\pm$10.36 & \textbf{87.03$\pm$13.16} \\
			& \makecell{MLP - Ordinal-Cross-Entropy Loss \\ + Class-Balancing} & 86.73$\pm$14.43 & 88.20$\pm$9.66 & 86.60$\pm$12.54 \\
			\hline
			\multirow{6}{*}{\makecell{B2\\ Temporal}} & ResNet - Cross-Entropy Loss & 84.75$\pm$13.21 & 83.10$\pm$11.92 & 82.72$\pm$12.74 \\
			& \makecell{ResNet - Cross-Entropy Loss \\ + Class-Balancing} & 84.03$\pm$15.13 & 83.28$\pm$11.42 & 82.97$\pm$12.84 \\
			& ResNet - Ordinal-Cross-Entropy Loss  & 85.24$\pm$15.68 & 87.23$\pm$10.52 & 85.56$\pm$13.38 \\
			& \makecell{ResNet - Ordinal-Cross-Entropy Loss \\ + Class-Balancing}  & 84.34$\pm$15.75 & 87.88$\pm$11.22 & \textbf{85.68}$\pm$\textbf{13.58} \\
			\hline
			
	\end{tabular}}
	
\end{table}

At first glance, the ResNet models perform slightly less than the MLP models. This could easily lead us to believe that simple histogram representations are enough to achieve a higher classification performance than the corresponding temporal representations. However, despite the performance differences, the temporal features and ResNet models help better explain and pin-point regions in the input feature space that contribute the most towards the model's decision. This is important if one wants to understand which student roles are most influential in the model's prediction. We will go over this aspect in more detail in the next section.

\begin{figure*}[tb!]
\centering
    \includegraphics[width=0.99\linewidth]{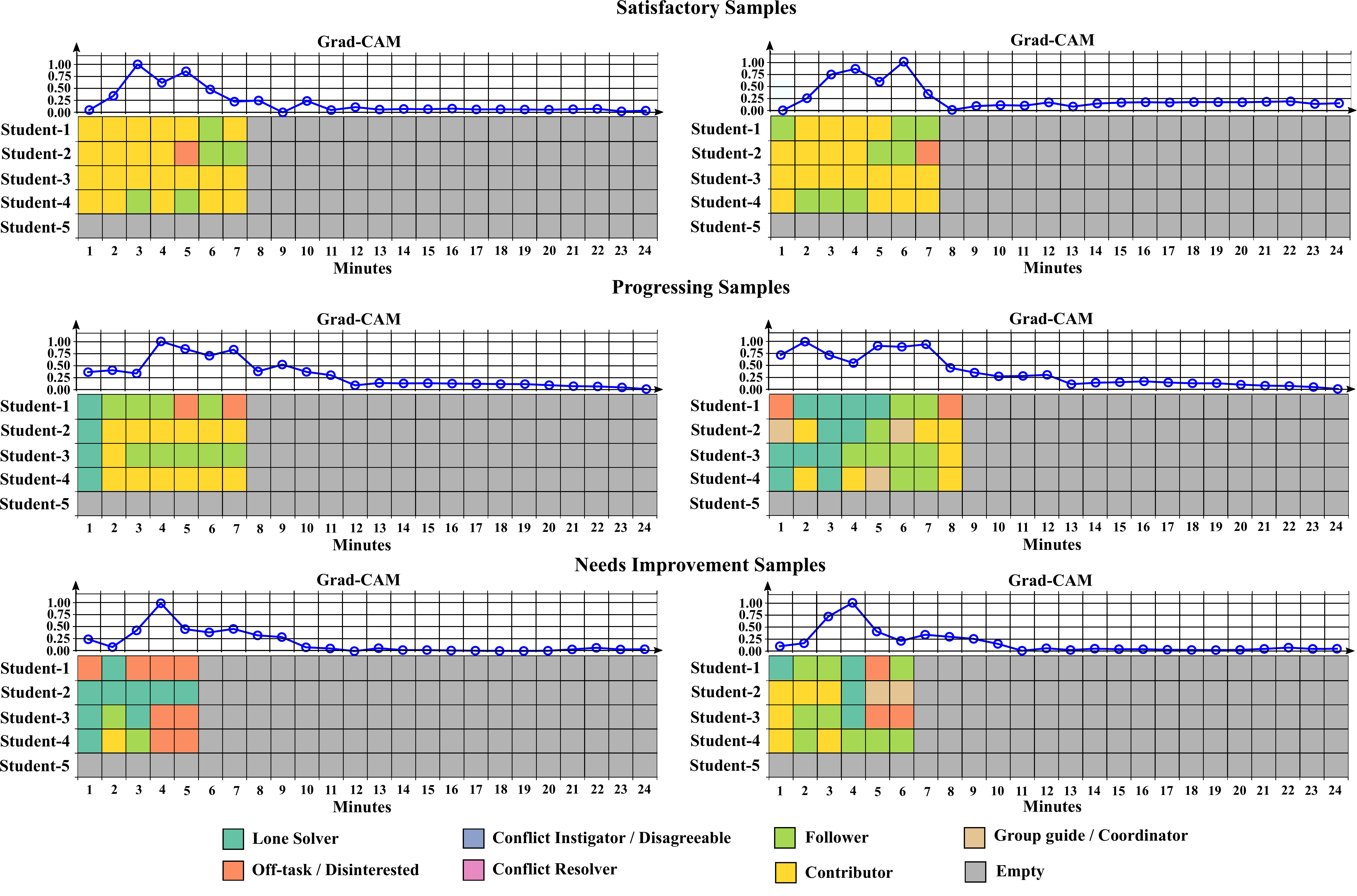}
	\caption{Grad-CAM visualization for two different temporal samples from different Level-A classes. 
	}\label{figure_GRADCAM_B2}
\end{figure*}

\subsection{Grad-CAM Visualization}

Grad-CAM uses class-specific gradient information, flowing into the final convolutional layer to produce a coarse localization map that highlights the important regions in the input feature space \cite{selvaraju2017grad}. It is primarily used as a post hoc analysis tool and is not used in any way to train the model. Figure \ref{figure_GRADCAM_B2} illustrates how Grad-CAM can be used for our classification problem. We show two different samples from the Satisfactory, Progressing and Needs Improvement classes respectively. Each sample shows a group consisting of 4 students that completed the task in 5 to 8 minutes. Technically one can obtain $C$ Grad-CAM maps for a $C$-class classification problem. Here, the samples shown correspond to the class predicted by the ResNet model, which is also the ground-truth class. It's clear how the Grad-CAM highlights regions in the input feature space that contributed towards the correct prediction. For instance, in the Needs Improvement examples, the Grad-CAM map shows the highest weight on the fourth minute. At that time for the first example, the codes for three of the students become Off-task/Disinterested. Similarly, for the second example we notice three of the students become Lone Solvers and the fourth student becomes a Follower. This is in stark contrast to the minute before when two of the students were Followers and the other two were Contributors. We also notice less importance being given to the Empty codes. These changes in roles and the Grad-CAM weights across the task make sense and help promote explainability in our deep learning models.

\section{Conclusion}\label{section_conclusion}

In this paper we proposed using simple temporal representations of individual student roles together with temporal ResNet deep-learning architectures for student group collaboration assessment. Our objective was to develop more explainable systems that allow one to understand which instances in the input feature space led to the deep-learning model's decision. We suggested use of Grad-CAM visualization along the temporal dimension to assist in locating important time instances in the task performed. We compared the performance of the proposed temporal representations against simpler histogram representations from our previous work \cite{som2020machine}. While histogram representations can help achieve high classification performance, they do not offer the same key insights that one can get using the temporal representations. 


\noindent \textbf{Limitations and Future Work:} The visualization tools and findings discussed in this paper can help guide and shape future work in this area. Having said that our approach can be further extended and improved in several ways. For example, we only discuss Grad-CAM maps along the temporal dimensions. This only allows us to identify important temporal instances of the task but does not focus on the important student interactions. The current setup does not tell us which subset of students are interacting and how that could affect the overall group dynamic and collaboration quality. To address this we intend on exploring other custom deep-learning architectures and feature representation spaces. We also plan on using other tools like LIME \cite{ribeiro2016should} and SHAP \cite{lundberg2017unified}. These packages compute the importance of the different input features and help towards better model explainability and interpretability. Also we only focused on mapping deep learning models from individual student roles to overall group collaboration. In the future we intend on exploring other branches in the conceptual model, described in \cite{alozie2020automated,alozie2020collaboration}. We also plan on developing recommendation systems that can assist and guide students to improve themselves by suggesting what they need to take on. The same system could also be tweaked specifically for teachers to give them insight on how different student interactions could be improved to facilitate better group collaboration.

\section{Acknowledgement}
This work was supported in part by NSF grant number 2016849.

%
\bibliographystyle{abbrv}
\bibliography{sigproc}  

\end{document}